\documentclass{article}
\usepackage{spconf,amsmath,graphicx}
\usepackage{mathtools}
\usepackage{amssymb}
\usepackage{accents}
\usepackage{hyperref}

\usepackage{enumitem}
\usepackage{gensymb}
\setlist{nosep, leftmargin=14pt}

\usepackage{longtable}

\title{Learning Spatially-Continuous Fiber Orientation Functions}
\name{Tyler Spears$^1$ \qquad P. Thomas Fletcher$^1$}

\address{$^1$ University of Virginia, Department of Electrical and Computer Engineering, Charlottesville, VA}

\begin{document}
%
\maketitle
\begin{abstract}
Our understanding of the human connectome is fundamentally limited by the resolution of diffusion MR images. Reconstructing a connectome's constituent neural pathways with tractography requires following a continuous field of fiber directions. Typically, this field is found with simple trilinear interpolation in low-resolution, noisy diffusion MRIs. However, trilinear interpolation struggles following fine-scale changes in low-quality data. Recent deep learning methods in super-resolving diffusion MRIs have focused on upsampling to a fixed spatial grid, but this does not satisfy tractography's need for a continuous field. In this work, we propose FENRI, a novel method that learns spatially-continuous fiber orientation density functions from low-resolution diffusion-weighted images. To quantify FENRI's capabilities in tractography, we also introduce an expanded simulated dataset built for evaluating deep-learning tractography models. We demonstrate that FENRI accurately predicts high-resolution fiber orientations from realistic low-quality data, and that FENRI-based tractography offers improved streamline reconstruction over the current use of trilinear interpolation. \footnote{Our project and code can be found at \url{https://osf.io/dvnxw/}}.
\end{abstract}
\section{Introduction}
\label{introduction}

Mapping the human connectome relies upon a continuous and accurate representation of the underlying brain tissue. This is needed for tracing streamlines, resolving crossing fibers, and deciding when to terminate a tract. Often, tractography algorithms rely on simple trilinear interpolation to ``fill out'' a continuous field from discretely-sampled diffusion magnetic resonance images (dMRIs). If this interpolation could be improved, then tractography algorithms could produce more detailed and accurate human white matter (WM) fiber tracts.

In this work, we propose FENRI (\textbf{F}iber orientations from \textbf{E}xplicit \textbf{N}eural \textbf{R}epresentat\textbf{I}ons), a novel deep learning-based super-resolution model for estimating fODFs continuously in space. We demonstrate FENRI's capabilities through the following experiments: 1) a quantitative evaluation of fODF reconstruction in Human Connectome Project (HCP) data, 2) a qualitative evaluation of tractography in HCP data, and 3) a quantitative measure of tractography performance on a new, expanded simulation dataset. As an image upsampler, FENRI outperforms more generic single-image super-resolution (SISR) methods on a variety of test metrics. We also show how, as a tractography enhancement, FENRI's explicit representation sampling provides a powerful improvement over standard tractography methods.

\textbf{Background.} Reconstructing streamlines from diffusion-weighted images (DWIs) requires a model of neuron fiber directionality. One popular model is the general fODF represented by coefficients in the spherical harmonic (SH) orthonormal basis, estimated by constrained spherical deconvolution (CSD) \cite{Jeurissen.etal.2014}. Several deep learning models have recently been proposed to super-sample diffusion representations. For example, Qin et. al., 2021 used convolutional neural networks (CNNs), an efficient sub-pixel CNN (ESPCN) layer, and high-resolution T1w volumes to predict high-resolution diffusion model parameters \cite{Qin.etal.2021,Shi.etal.2016}. However, these previous works were limited to upsampling by an \textit{integer upscaling factor}, e.g. $2 \times$, which is not ideal for estimating continuous fields. The recently proposed implicit neural representation (INR) method, which learns continuous-valued representations in some Euclidean space, is one solution to this challenge \cite{Sitzmann.etal.2020}. INRs are most commonly applied to 3D rendering, but INR-like models have been used in SISR. For example, the Local Implicit Image Function where a low-resolution input image is encoded into a feature space and sampled continuously for upsampling \cite{Chen.etal.2021}. To our knowledge, the only proposed model that utilizes INRs for super-resolving dMRIs is given in \cite{Consagra.etal.2023}, which focused on uncertainty in continuous predictions rather than tractography. We place FENRI alongside these INR models, but note that FENRI does not model an \textit{implicit} function, but an \textit{explicit} function of SH coefficients.

\section{Methods}
\label{methods}

Our objective is to \textbf{predict fODFs at arbitrary, continuous spatial coordinates} given only low-resolution DWIs. We have a set of subject DWIs $\mathbf{S}$ with a continuous space $\mathbf{\Omega} \subset \mathbb{R}^3$. 
Now, $\mathbf{S}$ is sampled on a discrete, finite, rectilinear grid $\mathbf{P} \subset \mathbf{\Omega}$ at coordinates $p \in \mathbf{P}$.
Given a \textit{query coordinate} $q \in \mathbf{\Omega}$, we wish to find the vector of SH coefficients $d_q$ that represents the fiber orientations at $q$. Thus, we construct a function $G_\theta$, with parameters $\theta$, such that $G_\theta(\mathbf{S}, \mathbf{P}, q) = d_q$.

\textbf{Encoder.} The encoder compresses the spatial and angular information found in the input DWIs, described as $E: \left(\mathbf{S}, \mathbf{P}\right) \rightarrow \mathbf{L}$, where $\mathbf{L}$ has feature vectors of length $c_{\mathbf{L}}$, and the spatial ordering of feature vectors in $\mathbf{L}$ is assumed to be equivalent to that of $\mathbf{P}$ (and $\mathbf{S}$). We chose a 3D implementation of the Cascading Residual Network (CARN) as the core of our encoder, a high-performing SISR architecture, \cite{Ahn.etal.2018}. We also add a kernel size 2 average-pooling layer and a single batch-norm layer at the output of the decoder to reduce checkerboard artifacts and improve training stability \cite{Sugawara.etal.2018}.

\textbf{Continuous Decoder.} The decoder predicts SH coefficients at any given query point $q \in \mathbf{\Omega}$ based on encoded features and the query coordinate itself, described by $D: (\mathbf{L}, \mathbf{P}, q) \rightarrow d_q$. The decoder at its core is a simple fully-connected network with $n_{D}$ hidden layers \cite{Chen.etal.2021}. Every query point $q$ requires 8 forward passes through the same decoder network. As mentioned in Chen et. al., 2021 as the ``local ensemble'', the features in $\mathbf{L}$ given to the decoder correspond to the $2 \times 2 \times 2$ spatially nearest-neighbors to $q$ as determined by the input DWI grid $\mathbf{P}$ \cite{Chen.etal.2021}. Each feature vector in this ensemble is given a forward pass through the decoder, and all $8$ outputs are trilinearly weighted as the final step in the model prediction. This is done to smooth the prediction when ``crossing over'' discrete points in $\mathbf{P}$ with $q$.

Here we describe a single forward pass of the decoder for a query $q$ and sub-grid point $\mathbf{L}_i \in \mathbf{L}$, the $i$'th input vector local to $q$, and its input coordinate $p_i$. The first layer takes a concatenation of $\mathbf{L}_i$, $p_i$, $q$, and the Fourier positional encoding of $p_i - q$. We normalize $p_i - q$ to $ [ 0, 1 ) $ (denoted $\text{N}_{\mathbf{P}}$) and use the encoding map $\mathcal{F}$ (with the number of frequencies $m$) proposed as the ``Positional encoding'' method in Tancik et. al., 2020 \cite{Tancik.etal.2020}. The full input size is $c_{\mathbf{L}} + 6 + 6m$, and the output is an SH coefficient vector.

\section{Experiments}
\label{experiments}

\begin{figure*}[tb!]
\centering
\includegraphics[width=\textwidth]{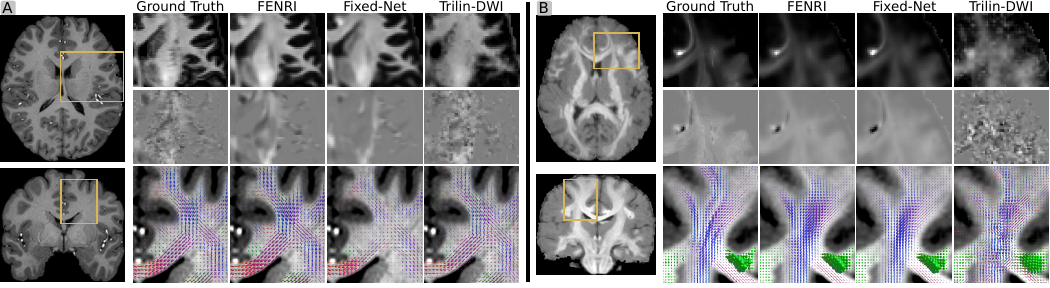}
\caption{Predicted SH coefficients on real HCP data and synthetic ISMRM-sim data. $(\textit{A})$ Prediction examples on an HCP subject, located at the yellow selection box on the subject's T1w image (left). Rows 1 and 2 are axial slices of SH indices of degree 0 order 0, and degree 6 order -1, respectively. Row 3 contains coronal slices of each methods's ODF lobe plot, with lobes scaled for visual clarity. $(\textit{B})$ Prediction examples on a synthetic ISMRM-sim subject. Image parameters match those in ($\textit{A})$.} 
\label{fig:vol_preds}
\end{figure*}

\textbf{Comparison Models.} We compare FENRI to a variety of upsampling and tractography methods. The baseline method, which we label as ``Trilin-DWI,'' is a trilinear upsampling of the noisy and low-resolution DWIs into the target spatial resolution. SH coefficients are then estimated from these upsampled DWIs with multi-shell, multi-tissue CSD (MSMT-CSD) \cite{Jeurissen.etal.2014}. We also evaluated a network that used the more common SISR ESPCN layer, which we named Fixed-Net (as in ``fixed-size upsampling network''). Fixed-Net utilized a similarly parameterized CARN encoder to that used by FENRI, allowing Fixed-Net to also serve as a rough FENRI ablation model. Fixed-Net's encoder (by way of its ESPCN layer) upsampled low-resolution DWIs by $2 \times$ the input spatial resolution, then used trilinear interpolation to resample its latent space into the target spatial resolution. The resampled latent space is then passed through a smaller CARN-style network to refine the trilinear resampling, producing an upsampled SH coefficient volume. For tractography, FENRI continuously samples its latent space $\mathbf{L}$ for SH coefficients at every tracking iteration, while all other models used trilinear sampling on predicted SH coefficients. The streamline tracking method was a deterministic gradient ascent-based tractography algorithm implemented for use on a graphics processing unit (GPU). We chose tracking parameters to closely match the defaults for the ``SD Stream'' tractography in MRtrix3 \cite{Tournier.etal.2019}.

\textbf{Experiment 1: HCP ODF Reconstruction.} We tested all models on voxel-wise ODF reconstruction of the HCP Young Adult dataset \cite{VanEssen.etal.2013}. All HCP DWIs were preprocessed with the standard HCP pipeline and normalized by b0 intensities \cite{Glasser.etal.2013,Tournier.etal.2019}. We degraded the DWIs to make them more clinically-realistic by: 1) angularly resampling gradient directions to the first 9 b0's, 45 b1000's and 45 b3000's of the HCP diffusion protocol, 2) downsampling from 1.25mm to 2.00mm, and 3) adding Rician-distributed noise to a 30 dB signal-to-noise ratio. Each model predicted a 1.25mm isotropic volume of even-degree SH coefficients with $l_{max} = 8$ from these degraded DWIs. Ground truth volumes were estimated from 1.25mm DWIs with multi-shell multi-tissue CSD (MSMT-CSD) and normalization \cite{Tournier.etal.2019,Jeurissen.etal.2014,Dhollander.etal.2021}. We evaluated models over three metrics: a weighted mean-squared error (WMSE) of the SH coefficients, the mean-squared Jensen-Shannon Distance (MSJSD) of the ODF, and the weighted average angular error (WAAE) of the ODF peaks. The WMSE is simple MSE with each degree $l$ scaled to a standard normal distribution; this is also the loss function for FENRI and Fixed-Net. The MSJSD calculates the JS Distance between the predicted and target ODFs with density functions estimated over a discrete set of directions \cite{Karimi.etal.2021}. Finally, the WAAE compares ground truth ODF peaks to the closest peak available in the predicted ODF \cite{Schultz.2012}. The angular distance was minimized between (at most) the two largest target peaks and (at most) the three largest prediction peaks, with a minimum peak of 0.1. False positive and negative prediction peaks were penalized by $0.0073 \pi / 2$, the empirically-estimated normalized median peak amplitude times the maximum angular distance.

\textbf{Experiment 2: HCP Qualitative Tractography.} We compared FENRI-based tractography with trilinear interpolation tractography on a real human subject. No method exists for directly measuring a ground truth tractogram in real, \textit{in vivo} HCP data, so our comparison was qualitative in nature. We input degraded DWIs (similarly to Experiment 1) to both FENRI and Trilin-DWI and bilaterally reconstructed two tract bundles, chosen for their recognition in the literature and differences in shape - the cortico-spinal tract (CST) and the uncinate fasciculus (UF).

\textbf{Experiment 3: ISMRM-sim Tractography.} In the recently updated ISMRM-2015 tractography challenge dataset, expert-curated WM bundles formed the ground truth of a simulated DWI dataset \cite{Renauld.etal.2023}. We extend the original dataset from one to 15 brain-like simulacrums to keep a train-test split for model training. We used symmetric image normalization registration (SyN)~\cite{Avants.etal.2008c} to match the original dataset's simulated T1w to 15 real HCP T1w images via the MNI-152 template, and warped the challenge's subjects WM bundles to each of the 15 target subjects. An example warped T1w image is in Figure \ref{fig:vol_preds}B. This allowed the dataset to maintain biological variability alongside the expert-selected ground truth bundles. Then, each target subject's warped bundles were used to create simulated 0.9 mm DWIs, with 4 b0's, 90 b1000's, and 90 b3000's (gradient directions match the standard HCP sequence) via Fiberfox simulation \cite{VanEssen.etal.2013,Neher.etal.2014}. These data are normalized and degraded similarly to the HCP DWIs in Experiment 1; we name this dataset ``ISMRM-sim.'' Thus, training FENRI and Fixed-Net on ISMRM-sim data requires upsampling of degraded 2.0 mm DWIs to 0.9 mm SH coefficients. Examples of ISMRM-sim SH coefficients are shown in Figure \ref{fig:vol_preds}B. We aimed to measure each model's ability to reconstruct the bundles used to create the ISMRM-sim DWIs. Tractography was seeded at the length-wise midpoint of each ground truth streamline in the following bundles: brainstem projection system (BPS), corpus callosum temporal (CC-t), corpus callosum u-shaped (CC-u), cingulum bundle (Cing), optic radiation (OR), inferior longitudinal fasciculus (ILF), superior longitudinal fasciculus (SLF), and uncinate fasciculus (UF). Performance was evaluated with the challenge's tractogram rating script on each separate bundle, giving the same metrics as shown in the challenge results: overlap voxel ratio (OL), overreach voxel ratio normalized by bundle volume (OR), and segmentation f1 (Dice) score.

\section{Results \& Discussion}
\label{results}

\begin{figure}[tb!]
\centering
\includegraphics[width=\linewidth]{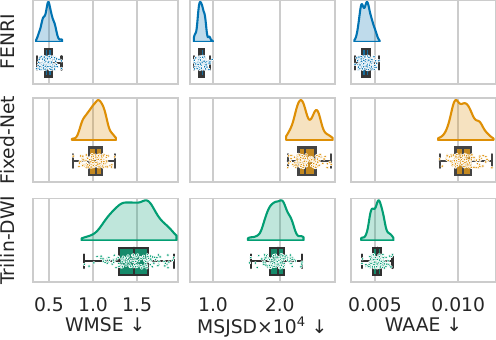}
\caption{Results of predicting HCP SH coefficients from low-resolution DWIs in white matter voxels. Each point is a score for a single HCP subject; the distribution plots summarize the point data. Arrows indicate the direction of better performance. WMSE: weighted mean squared error; MSJSD: mean-squared Jensen-Shannon distance; WAAE: weighted average angular error.}
\label{fig:hcp_vox_results_plot}
\end{figure}

\textbf{Experiment 1: HCP ODF Reconstruction.} The quantitative results for predicting HCP SH coefficients are given in Figure \ref{fig:hcp_vox_results_plot}, and example predictions are illustrated in Figure \ref{fig:vol_preds}A. Across all three metrics (WMSE, MSJSD, and WAAE), FENRI outperformed both Trilinear-DWI and Fixed-Net, while also showing equal or lower variance between test set subjects. As shown in Figure \ref{fig:vol_preds}A, FENRI gives high-quality predictions that preserve high frequency details, both in the spatial and angular sense. Against Fixed-Net, FENRI better reconstructs the high degree SH coefficients and produces an overall more accurate ODF, despite sharing an encoder architecture with Fixed-Net. While Trilin-DWI is impressive as a baseline, FENRI still gives quantitatively better predictions. One shortcoming of Trilin-DWI can be seen at the gray matter-WM boundary in Figure \ref{fig:vol_preds}A, row 3, where Trilin-DWI cannot maintain edges over sharp turns in the gyri; FENRI, however, better preserves these boundaries. Additionally, Fixed-Net's performance is surprising compared to Trilin-DWI. Fixed-Net outperformed Trilin-DWI on WMSE, the network's optimized loss function. However, Trilin-DWI outperforms Fixed-Net on both MSJSD and WAAE. We hypothesize that this is caused by Fixed-Net overfitting to the objective function, while struggling to reconstruct high-degree SH coefficients on the real HCP data. This can be seen in Figure \ref{fig:vol_preds}A, where Fixed-Net's predicted degree 6 coefficient is relatively sparse. However, this weakness of Fixed-Net does not seem as limiting in the more homogeneous ISMRM-sim dataset, as shown in Figure \ref{fig:vol_preds}B. Overall, these Fixed-Net results suggest that more common SISR methods may not be appropriate for predicting and upsampling ODFs on real data.

\begin{figure}[tb!]
\centering
\includegraphics[width=\linewidth]{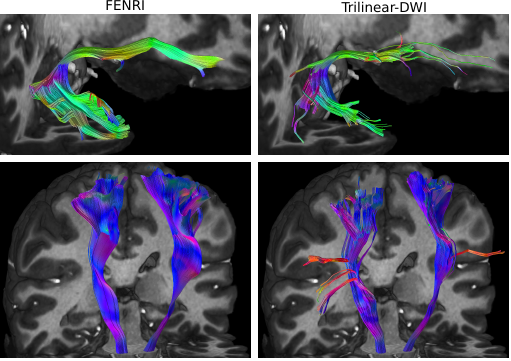}
\caption{Qualitative results of both trilinear and FENRI tractography on an HCP subject between FENRI and Trilin-DWI. Row 1 shows an example unilateral uncinate fasciculus, and row 2 shows the left and right cortico-spinal tracts.}
\label{fig:hcp_tract_results}
\end{figure}

\textbf{Experiment 2: HCP Qualitative Tractography.} The qualitative results of both trilinear and FENRI tractography are shown in Figure \ref{fig:hcp_tract_results}. Tracking was poor when interpolating on Fixed-Net predictions of HCP data and were omitted for brevity. This poor tracking becomes clear when seeing the Fixed-Net predicted lobe plots shown in the final row of Figure \ref{fig:vol_preds}A. When comparing FENRI tractography to trilinear DWI upsampling+tractography, FENRI generally produces smoother, more filled tracts when given the same seeds and tracking parameters. Row 1 in Figure \ref{fig:hcp_tract_results} illustrates an example UF where FENRI-based tractography produces a smooth and dense tractogram when compared to Trilin-DWI. A similar result is found in reconstructed CST. Seed points were only given in the mid-Pons area of the brainstem, so all produced tracts must nearly traverse the entire brain longitudinally, maximizing the chance of integration errors while tracking.

\begin{figure}[tb!]
\centering
\includegraphics[width=\linewidth]{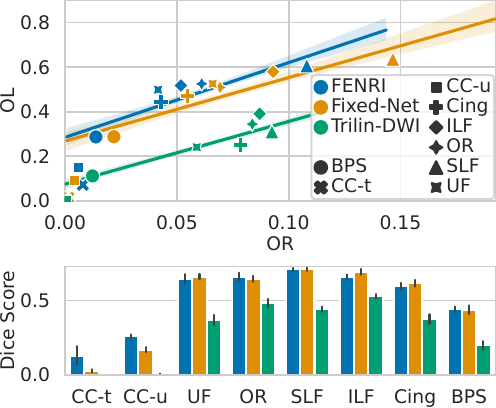}
\caption{Quantitative tractography results on ISMRM-sim. \textit{(Top)} overreach (OR) vs. overlap (OL) of each tractography method over selected bundles. Points represent method and bundle score averaged over all subjects. Regression lines are fit to each method's scores, shaded on the $95\%$ confidence interval. \textit{(Bottom)} Dice scores for each method and bundle. Colors match the top row, and error bars represent the $95\%$ confidence interval. See Section \ref{experiments} for the full bundle names.}
\label{fig:ismrm_sim_results}
\end{figure}

\textbf{Experiment 3: ISMRM-sim Tractography.} Figure \ref{fig:ismrm_sim_results} shows the quantitative results of ISMRM-sim data tractography. Figure \ref{fig:vol_preds}B illustrates that Trilin-DWI is challenged by the sparse, thin features of the simulation. These thin streamlines are obscured by downsampling and noise injection, so tractography becomes difficult without learned priors, such as those learned by FENRI and Fixed-Net. Comparing the learned networks, FENRI matches or exceeds Fixed-Net in most given metrics. As shown in Figure \ref{fig:ismrm_sim_results}A, FENRI produces an equal or higher OL on all tested bundles, while having a generally lower OR, particularly in the SLF, ILF, and the Cingulum. In Figure \ref{fig:ismrm_sim_results}B, FENRI typically matches or outperforms Fixed-Net's Dice score, especially on the CC bundles. However, Fixed-Net does outperform FENRI in the ILF, and matches FENRI in most other bundles. We hypothesize that the selected bundles are traversable by trilinear interpolation given a sufficiently high-resolution ODF. This warrants further analysis on other bundles, and gives opportunities for improving FENRI. Overall, these results suggest that using FENRI in tractography can produce more accurate tractograms, particularly in smaller WM bundles.

\textbf{Discussion.} We have proposed and evaluated FENRI, a first-of-its-kind explicit neural representation model built for tractography. We have shown that FENRI performs well in predicting and upsampling fODFs on a continuum of spatial resolutions when given degraded, clinical-quality DWIs. We have also shown that FENRI can be effectively used in tractography and holds great potential for improving a variety of tractography-focused methods. Finally, we have expanded the ISMRM 2015 challenge dataset to 15 HCP subjects and made these data publicly available for future works in quantitative evaluation of deep learning models in tractography.


\textbf{Compliance with Ethical Standards} This research was conducted using the publicly released HCP Young Adult human dataset. Ethical approval was not required.

\textbf{Acknowledgments} Data were provided in part by the Human Connectome Project, WU-Minn Consortium (Principal Investigators: David Van Essen and Kamil Ugurbil; 1U54MH091657) funded by the 16 NIH Institutes and Centers that support the NIH Blueprint for Neuroscience Research; and by the McDonnell Center for Systems Neuroscience at Washington University.

\bibliographystyle{IEEEbib}
\bibliography{strings,refs}

\end{document}